\documentclass[runningheads]{llncs}
\usepackage[T1]{fontenc}
\usepackage[utf8]{inputenc}
\usepackage{graphicx,verbatim}
\usepackage{amssymb, booktabs, xcolor, caption, siunitx, makecell, hyperref, listings, multirow}
\usepackage{listings}
\usepackage[utf8]{inputenc}
\usepackage[export]{adjustbox}
\usepackage{lipsum}
\newcommand{\TitleLogo}{%
\raisebox{0ex}{
  \includegraphics[valign=c,width=0.12\textwidth]{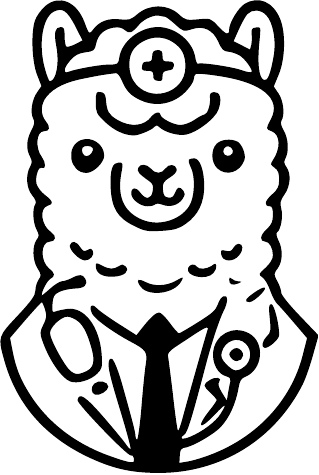}
  }
}
\newcommand{\FullTitle}{%
  \parbox[c]{0.739\textwidth}{
    $\mu^{2}$Tokenizer: Differentiable\\[-0.2ex]
    Multi-Scale Multi-Modal\\[-0.2ex]
    Tokenizer for\\[-0.2ex]
    Radiology Report Generation
    }
}
\begin{document}
\title{%
     \TitleLogo\hspace{1em}\FullTitle
}

\author{
  Siyou Li\inst{1}
 \and Pengyao Qin\inst{2}
 \and Huanan Wu \inst{3}
 \and Dong Nie\inst{4}
 \and Arun J. Thirunavukarasu\inst{5}
 \and Juntao Yu\inst{1}
 \and Le Zhang\inst{2,6}
}

\authorrunning{Siyou Li et al.}
\titlerunning{$\mu^2$Tokenizer: Differentiable Multi-Scale Multi-Modal Tokenizer}
\institute{School of Electronic Engineering and Computer Science, \\Queen Mary University of London, London, UK \\
\and 
School of Engineering, College of Engineering and Physical Sciences, \\University of Birmingham, Birmingham, UK\\
\and 
Guangdong University of Technology, Guangdong, China\\
\and 
Meta Inc. US\\
\and 
Nuffield Department of Clinical Neurosciences,\\ University of Oxford, Oxford, UK\\
\and 
William Harvey Research Institute, NIHR Barts Biomedical Research Centre, Queen Mary University London, London, UK\\
\email{\{siyou.li, juntao.yu\}@qmul.ac.uk; l.zhang.16@bham.ac.uk}
}
 
\maketitle        
\begin{abstract}
Automated radiology report generation (RRG) aims to produce detailed textual reports from clinical imaging, such as computed tomography (CT) scans, to improve the accuracy and efficiency of diagnosis and provision of management advice. RRG is complicated by two key challenges: (1) inherent complexity in extracting relevant information from imaging data under resource constraints, and (2) difficulty in objectively evaluating discrepancies between model-generated and expert-written reports. To address these challenges, we propose $\mu^2$LLM, a $\underline{\textbf{mu}}$ltiscale $\underline{\textbf{mu}}$ltimodal large language models for RRG tasks. The novel ${\mu}^2$Tokenizer, as an intermediate layer, integrates multi-modal features from the multiscale visual tokenizer and the text tokenizer, then enhances report generation quality through direct preference optimization (DPO), guided by GREEN-RedLlama. Experimental results on four large CT image-report medical datasets demonstrate that our method outperforms existing approaches, highlighting the potential of our fine-tuned $\mu^2$LLMs on limited data for RRG tasks. At the same time, for prompt engineering, we introduce a five-stage, LLM-driven pipeline that converts routine CT reports into paired visual-question-answer triples and citation-linked reasoning narratives, creating a scalable, high-quality supervisory corpus for explainable multimodal radiology LLM.
All code, datasets, and models will be publicly available in our official repository.\footnote{\url{https://github.com/Siyou-Li/u2Tokenizer}}


\end{abstract}

\section{Introduction}

Radiology reports are the primary means by which radiologists communicate their findings, likely diagnoses, and management recommendations to referring physicians and surgeons~\cite{zhao2023radiology}. These reports must be accurate and interpretable, as ambiguous language or mistakes can lead to clinical error as well as increased patient anxiety~\cite{rosenkrantz_differences_2017}. Expert reports are especially important for imaging that referrers are frequently unable to interpret independently, such as computed tomography (CT). 
An increasing volume of CT examinations year-on-year generates pressure for radiologists to produce more high-quality reports, compounded by workforce shortages~\cite{everlight_radiology_2025}. Emerging artificial intelligence (AI) and natural language processing (NLP) technologies built upon foundation model architectures show promise in automating radiology report generation (RRG)~\cite{thirunavukarasu_large_2023}. If accurate, automated RRG may streamline radiologist workflows, reduce reporting time, and enhance report quality. Automated RRG may also facilitate large-scale data extraction for clinical research, improving the usability of radiological data~\cite{ostmeier2024greengenerativeradiologyreport}. Integrating AI into clinical practice may thereby enhance diagnostic accuracy, improve patient outcomes, and help meet the demand for healthcare services.

\begin{figure}[!t]
    \centering
    \includegraphics[width=\textwidth]{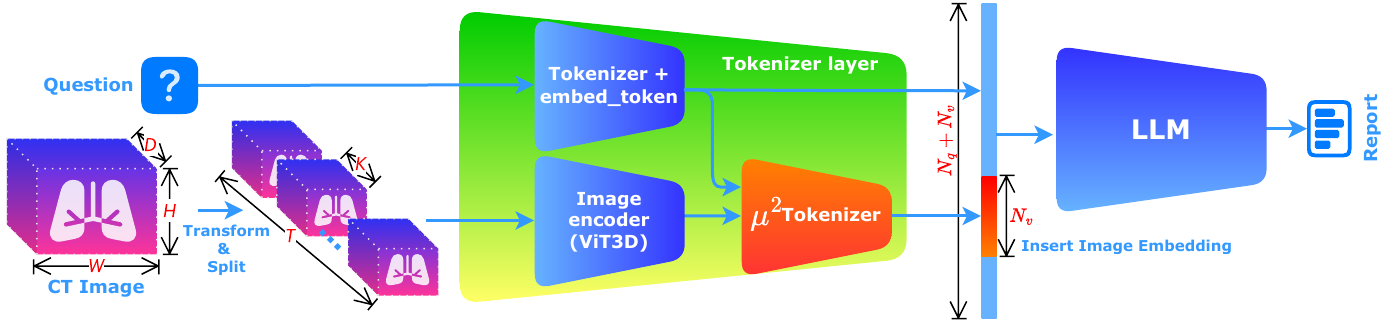}
    \caption{Overview of our proposed $\mu^2$LLM model that is centered with the $\mu^2$Tokenizer layer for high quality RRG task.}
    \label{fig:pipeline}
\end{figure}

Existing RRG models are typically built around LLaVA~\cite{liu2023llava}, where input data consists of CT images resized to fixed dimensions. However, CT images exhibit variability in length, width, and height, and the resizing processes can distort anatomical details and lesions, potentially compromising diagnostic accuracy. Moreover, the direct incorporation of high-resolution CT images into analytical pipelines is inefficient and frequently prohibited by limited computational resources, rendering the comprehensive and efficient extraction of pertinent imaging data a critical challenge in the report generation. Additionally, there is no unified standard for structuring radiology reports; clinicians prioritize the content of the findings over strict character-level alignment. Traditional NLP evaluation metrics, such as BLEU~\cite{papineni-etal-2002-bleu} and ROUGE~\cite{lin-2004-rouge}, are therefore not well-suited for evaluating RRG because they focus on lexical similarity rather than clinical salience or meaning. By comparing mainstream RRG models~\cite{wu2023generalist},~\cite{bai2024m3d}, we identify two significant challenges: (1) limitations in image encoders, as conventional approaches that crop or downsample CT scans lead to significant information loss, particularly along organ boundaries; and (2) the inadequacy of conventional NLP evaluation methods, which fail to accurately capture the semantic and clinical relevance of generated reports compared to ground truth. 
\textbf{Our contributions:} we propose a novel automated RRG approach based on a multi-modal large language model (MLLM) named $\mu^2$LLM, as shown in Fig.~\ref{fig:pipeline}. Our framework is designed to efficiently and cost-effectively preserve critical imaging details through guided question integration within the MLLM. Central to our approach is the proposed $\mu^2$Tokenizer, an intermediate processing layer that applies multi-level attention mechanisms and multi-scale aggregation on the outputs of the visual tokenizer (ViT3D). This layer further incorporates multi-modal attention to seamlessly fuse input question embeddings with refined CT image embeddings, thereby maximizing their semantic correspondence while maintaining computational efficiency. For evaluation, we identify and elucidate clinically significant errors using the GREEN model~\cite{ostmeier2024greengenerativeradiologyreport}—a specialized RRG metric that leverages large language model-based natural language understanding. To enhance the quality of generated reports, we employ direct preference optimization (DPO)~\cite{rafailov2023direct} to align model outputs with expert-validated clinical accuracy. Comprehensive evaluations conducted on three extensive CT image-report datasets demonstrate that our method produces radiology reports that contain clinically salient points and are computationally efficient, addressing critical limitations exhibited by existing automated RRG models.

\section{Method}

\subsection{Problem Set-up}

Given a CT image $\textbf{I}$ and the corresponding question text $\textbf{Q}$, the RRG task aims to generate a report  $\hat{\mathbf{Y}}$ that describes diagnostic findings $\mathbf{Y}$ by answering the given question $\textbf{Q}$. 

The perceiver approaches~\cite{jaegle2021perceiver} do not consider the $\textbf{Q}$ when compressing the image embeddings, which leads to suboptimal solutions. 
In our work, we introduce an intermediate tokenization layer, $\mu^2$Tokenizer, to effectively bridge the vision and language models. Figure~\ref{fig:pipeline} shows the pipeline of our model.

\textbf{Image Encoder and Text Tokenizer:} 
We scale the CT image $\textbf{I}$ with height $H$, width $W$, and depth $D$, and then divide it into $T$ frames each consisting of $K$ slices. i.e. $\textbf{I} \in \mathbb{R}^{T \times K \times H \times W}, T=\frac{D}{K}$. we employ a Vision Transformer (ViT3D)~\cite{bai2024m3d} as our image encoder, which first transforms and splits each frame $I_i, i \in[0, T)$ 
into frames and patches. The purpose of splitting the CT into several frames is to reduce the huge amount of computation caused by one-time input, and to avoid the information loss caused by direct downsampling and splitting. The image encoder then extracts local features for each patch and a global representation of the CT image. Formally, this produces a sequence of visual tokens: $\mathbf{V} \in \mathbb{R}^{T \times N_v \times E}$, where  $N_v$  is the number of visual tokens, and $E$ is the embedding dimension. Simultaneously, we obtain text tokens $\mathbf{Q}\in \mathbb{R}^{N_q \times E}$ after tokenizing the question with the text tokenizer, where $N_q$ is the max length of the question.
The $\mu^2$Tokenizer fuses the text tokens $\mathbf{Q}$ with visual tokens  $\mathbf{V}$ to create compact visual tokens $\mathbf{V'} \in \mathbb{R}^{N_v \times E}$ using a multi-scale multi-modal attention mechanism. 
This ensures that relevant image information is efficiently passed to the LLM while reducing computational overhead.

\textbf{Report Generation:} The processed image embeddings are then integrated with a text question to generate a radiology report. We utilize M3D-LaMed-Phi-3-4B~\cite{bai2024m3d} as the base LLM for report decoding. The LLM takes as input both the textual question $\mathbf{Q}$ and the $\mu^2$Tokenizer-processed visual tokens $\mathbf{V'}$, generating the radiology report $\hat{\mathbf{Y}}$ accordingly.

\subsection{$\mu^2$Tokenizer: Differentiable Multi-Scale Multi-Modal Tokenizer}
\begin{figure}[!t]
    \centering
    \includegraphics[width=\textwidth]{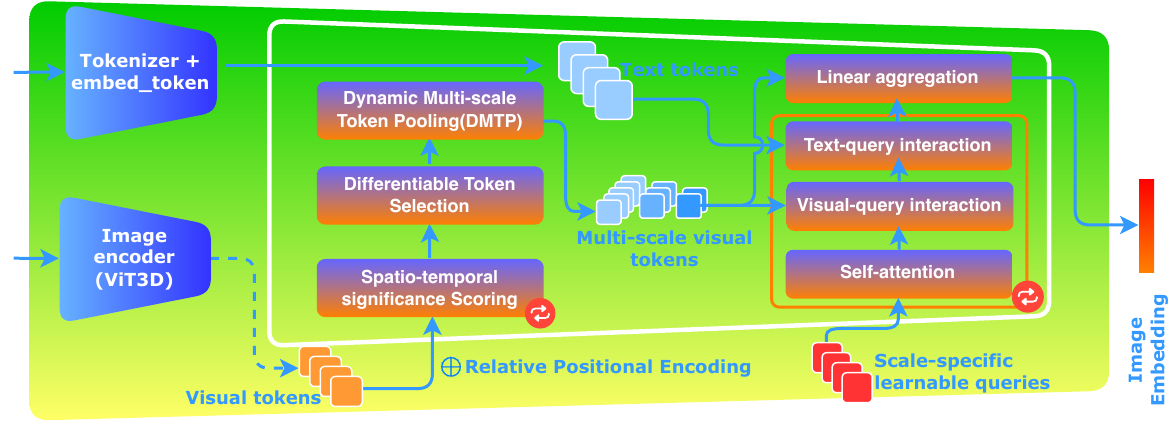}
    \caption{The illustration of our proposed $\mu^2$Tokenizer. The improvement is applied to steps of Token Selection, Multi-scale Pooling, and the Positional Encoding.}
    \label{fig:linear3dtokenizer}
\end{figure}

To improve the extraction efficiency for the CT image information, we propose the $\mu^2$Tokenizer module (as shown in Figure~\ref{fig:linear3dtokenizer}), which can process CT images with an arbitrary number of slices and leverage the pre-trained model for efficient alignment training. This module is built upon the Linear Video Tokenizer (LinVT)~\cite{gao2024linvt} that was originally introduced for the video understanding task. LinVT comprises two sub-modules: Spatio-temporal Visual Token Refiner (SVR) and Text-conditioned Token Aggregation (TTA). These modules adhere to the linearity principle, meaning that the output of each module is a linear combination of part of its input, thereby preserving the visual-language alignment learned in the image-LLM. 

\textbf{Relative Positional Encoding(RPE).} The LinVT~\cite{gao2024linvt} uses absolute learnable positional embeddings added along the frame and token dimensions. For the $j_{th}$ visual token $V_{i,j}$ in $i_{th}$ frame, two absolute positional embeddings ($P_f(i)$ and $P_t(j)$) are added to the $V_{i,j}$. Such an approach is not effective in capturing local relationships that are particularly useful in 3D volumes where local patterns matter. 
Instead, we use relative positional encoding so that the model can better capture local relationships regardless of the absolute position. 
The relative positional encoding is integrated within the attention mechanism\cite{shaw2018selfattentionrelativepositionrepresentations}. 
When computing the attention score between token $i$ and token $j$, we add learned positional embeddings based on their relative position: $A_{ij} = \frac{Q_i K_j^\top}{\sqrt{d}} + P_r(i-j)$. When used with multi-head attention, each head has its unique positional embeddings. 

\textbf{Differentiable Token Selection(DTS).} The LinVT~\cite{gao2024linvt} uses a hard top‑k selection which results in information loss when visual tokens are not selected. From a training perspective, a small $k$ also leads to slow optimization as the error can not be effectively backpropagated back to non-selected visual tokens. To solve this limitation, we replace the hard selection with a fully differentiable soft selection mechanism. For each of the $k$ selections, it computes a weight for all tokens and uses the weighted sum to produce a “soft” top token. This not only mitigates the information loss but also makes the selection process fully differentiable and improves gradient flow. The top-k soft tokens were calculated globally, taking into account visual tokens in all frames. Formally, for each of the $k$ soft tokens we first compute an attention score $\alpha^{(r)} = Softmax(W_s^{(r)}\,V_{flat}) \text{ where } W_s^{(r)}\in \mathbb{R}^{E \times 1}, V_{flat}\in \mathbb{R}^{T\cdot N_v\times E}, r = 1, \dots, k$. The soft token $V_{top}^{(r)}$ is calculated as the weighted sum of all visual tokens: 
$V^{(r)}_{top} = \sum_{i=1}^{T\cdot N_v} \alpha_{i}^{(r)} \, V_{flat}(i)$.

\textbf{Dynamic Multi-scale Pooling (DMTP).} The LinVT uses fixed pooling kernel sizes that treat individual kernels equally. We improved it to dynamic pooling, which allows the network to learn how to weight and choose the appropriate pooling. This is a more effective alternative to fixed pooling. We adapt the dynamic multi-scale pooling to weight the multiple pooled outputs dynamically. To implement this, we first apply an average pooling on different kernel sizes $s \in [1,2,4]$:
$
y_s = \text{AvgPool}(V_{top}, kernel=s)
$
and then compute dynamic weights $w_s$ via a small MLP $g(\cdot)$ over the mean of the pooled outputs:
$
w_s = \frac{\exp(g(\overline{y}_s))}{\sum_{s{\prime} \in S} \exp(g(\overline{y}_{s{\prime}}))}.
$
The weight $w_s$ is multiplied by $y_s$ to create weighted pooling outputs.
Then, the final pooled representation is created by concatenating these weighted outputs. This allows the network to adapt the pooling operation based on the input distribution.

\subsection{Direct Preference Optimization with GREEN-Score}

\begin{figure*}[!t]
    \centering
    \includegraphics[width=1\textwidth]{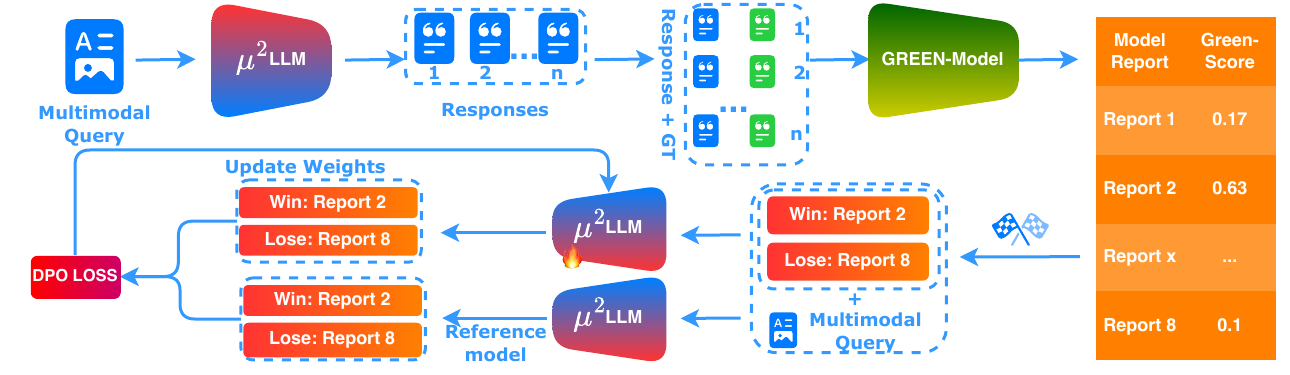}
    \caption{Overview of training process with Direct Preference Optimization (DPO).}
    \label{fig:data_preprocess}
\end{figure*}
Our training consists of two stages. First, we perform supervised fine-tuning (SFT) using the CT-Reports dataset. Second, we adopted the Direct Preference Optimization(DPO)~\cite{rafailov2023direct} method (as shown in Figure~\ref{fig:data_preprocess}). In particular, we optimize our model on GREEN Score~\cite{ostmeier2024greengenerativeradiologyreport}, which is arguably the most effective method to evaluate medical reports. GREEN~\cite{ostmeier2024greengenerativeradiologyreport} effectively identifies significant discrepancies between the reference and generated reports, providing a detailed score from 0 to 1 for quantitative analysis and a summary for qualitative analysis. This interpretable evaluation helps improve the quality of automated radiology reporting.
To obtain the preference dataset, we use the trained SFT model to generate a large number of medical reports on the existing dataset, and then these medical reports are scored by GREEN against the ground truth. Finally, the scored reports are used in DPO~\cite{rafailov2023direct} training, to guide the model generating preferred reports that have the highest GREEN score. As a result, the reports generated by our model are more accurate and semantically closer to human experts. Formally, the model is trained on the following \textbf{DPO training objective}:
$$L_{DPO}(\pi_\theta;\pi_{ref})=-\mathbb{E}_{(V,x,y_w,y_l)\sim \mathcal{D}_{DPO}} [\log{\sigma}(\beta \log{\frac{\pi_\theta(y_w | x)}{\pi_{ref}(y_w|x)}-\beta\frac{\pi_\theta(y_l|x)}{\pi_{ref}(y_l | x)}})]$$
with $\pi_\theta$ is the policy model, $\pi_{ref}$ is reference model, $\sigma$ is sigmoid function, $\beta \in (0.1,0.5)$, $y_w$ represent good responses and $y_l$ represent bad responses.

\subsection{Prompt Engineering}
\begin{figure}[]
    \centering
    \includegraphics[width=\textwidth]{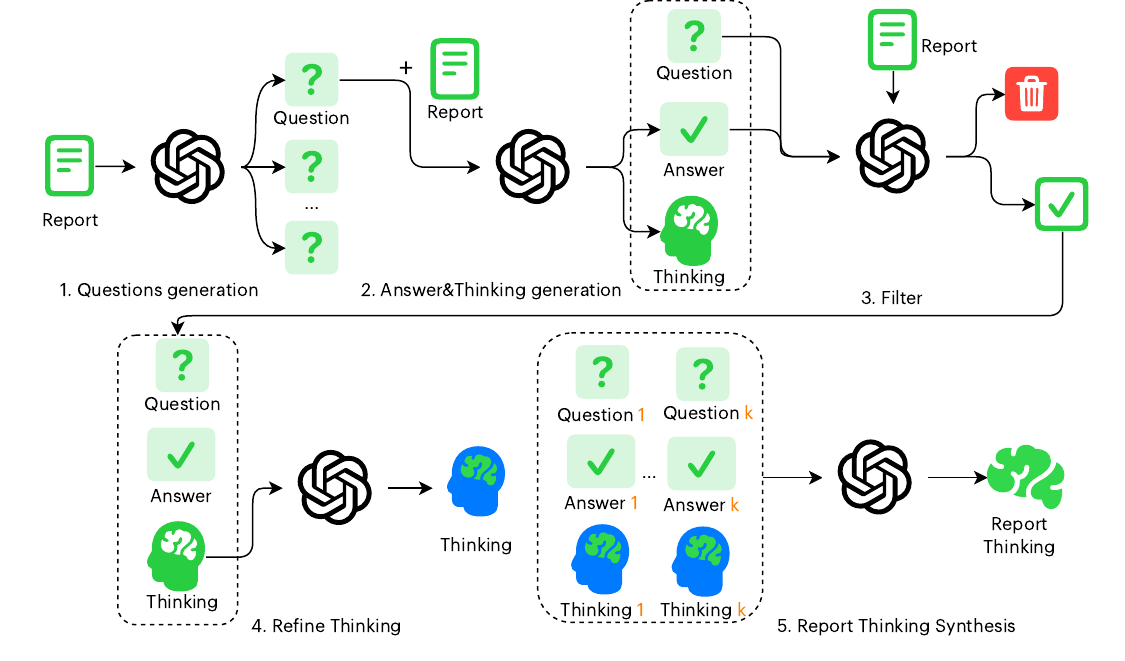}
    \caption{The pipeline of our proposed CT Report Reasoning Synthesis}
    \label{fig:prompt_engineering}
\end{figure}
In our workflow, we applied three prompt-engineering techniques—CT Report Rewriting, CT Report Reasoning Synthesis, and CT Report Translation—with particular emphasis on CT Report Reasoning Synthesis. The concrete process and involved prompts are listed in Appendix A.

Our CT Report Reasoning Synthesis pipeline converts each CT Report and its free-text radiology report into a rich supervisory package for multimodal learning by sequentially prompting a single large-language model in five roles: 

First, the question-generation stage reads the full report (findings and impression) and asks the LLM to propose a diverse collection of natural-language questions that a radiologist, trainee or downstream AI system might reasonably ask. Prompt constraints force coverage across lesion attributes, anatomical localisation, diagnostic certainty, and suggested follow-up, giving each study a rich inquiry space.

Second, each question is paired with the original report and resubmitted to the LLM under a “think-step-by-step” instruction.  The model must clearly reason out, citing exact report fragments or well-established imaging priors, before providing a concise answer.  The resulting tuples—question, answer, and raw reasoning—capture both knowledge and justification in a single pass.

Third, an automatic quality gate re-examines every tuple.  A second LLM pass checks factual consistency between answer and report, heuristics reject non-English or vacuous chains-of-thought, and domain-specific rules eliminate pathophysiologic contradictions (for example, claiming a pneumothorax is “improved” when it is first detected).  Only tuples that survive all three filters remain.

Fourth, accepted reasoning traces are refined: the LLM compresses them into short, evidence-linked paragraphs whose citations reference specific report lines.  Redundancy is pruned, hedging language is toned down and, where appropriate, probabilistic qualifiers are inserted to reflect clinical uncertainty in a calibrated fashion.

Finally, the pipeline fuses all refined traces into a single, structured “report-thinking” narrative.  The LLM merges overlapping rationale, orders arguments anatomically and separates them into Findings Rationale, Impression Rationale and Follow-up Rationale sections.  The finished datapoint therefore contains a CT volume, its VQA pairs (with answers) and a coherent explanation grounding every key statement, enabling scalable training of multimodal models that can answer questions and justify their answers with radiologic evidence.

\section{Experiments}

\textbf{Datasets.} \textbf{AMOS-MM 2024}~\cite{ji2022amos} consists of 2,088 chest, abdomen, and pelvis medical images and corresponding manually annotated text reports. The medical images are CT scans with spatial resolution from 256$\times$256 to 1024$\times$1024 and slice thickness from 1$mm$ to 5$mm$. 
\textbf{CT-Rate}~\cite{hamamci2024foundation} consists of 50,188 CT images of 21,340 patients and corresponding text reports. 
The scanning resolution and slice numbers range from 256$\times$256 to 1024$\times$1024 and 46 to 1,277, respectively. \textbf{AbdomenAltas 3.0}~\cite{bassi2025radgpt} is created by using RadGPT to generate reports for 17 public datasets, and through annotation, review, and refinement by 12 radiologists to ensure the reports' accuracy.
It comprises over 1.8 million text tokens and 2.7 million images from 9,262 CT scans. 



Furthermore, we expanded the dataset based on manually annotated medical reports using GPT-4o mini. This expansion included report rewriting and the generation of clinically relevant question-answer pairs, enriching the dataset’s diversity and comprehensiveness for improved model training and evaluation.

\textbf{Implementation Details.} We preprocess the 3D CT images using Min-Max Normalization, then resizing and cropping to a standard dimension of $8 \times 32 \times 256 \times 256$, and a random noise added. Our 3D vision encoder employs a 3D ViT from M3D-CLIP~\cite{bai2024m3d}, and base-LLM is Llama-3.2-1B-Instruct~\cite{touvron2023llamaopenefficientfoundation}.
All models are trained by AdamW optimizer with warm-up and cosine decay and use the bf16 mixed-precision training strategy enabled by DeepSpeed. Training is conducted in parallel across 4 NVIDIA A40 GPUs (48 GB VRAM each). For the $\mu^2$Tokenizer layer, we use four Spatio Temporal Attention Layers and four Text Condition Token Attention layers each consisting of eight attention heads and set $k = 1024$ for top soft token selection. For scale-specific learnable queries we use 1024 queries with a hidden size of 768.

\textbf{Baselines and Evaluation Metrics.} We compare our model with several efficient and high-performing MLLMs, including LaMed-Phi-3-4B~\cite{bai2024m3d}, LaMed-Llama-2-7B~\cite{bai2024m3d}, CT-CHAT(Llama-3.1-8B)\cite{hamamci2024developinggeneralistfoundationmodels}, RadGPT-N~\cite{bassi2025radgpt}, and RadFM-14B~\cite{wu2023generalist}, which excel at capturing linguistic patterns and generating coherent text across various domains. We also include the comparison of our model $\mu^2$LLM-1B (SFT) with only SFT, and our model $\mu^2$LLM-1B (SFT\&DPO) with both SFT and DPO.
Given the complexity of evaluating content accuracy between generated reports and human references, we employ both traditional and LLM-based metrics. Traditional metrics include BLEU~\cite{papineni-etal-2002-bleu}, ROUGE~\cite{lin-2004-rouge}, METEOR~\cite{banerjee-lavie-2005-meteor}, and BERT-Score\cite{bert_score}, which quantify text similarity through n-gram overlap and variations, although they have limited semantic understanding. LLM-based metrics, i.e. the GREEN score~\cite{ostmeier2024greengenerativeradiologyreport}, utilize models with strong semantic comprehension to evaluate the alignment between generated reports and human references. This metric assesses matching content and errors, offering a more comprehensive report quality measure.

\begin{table}[!t]
\centering
\setlength{\tabcolsep}{1mm}
\renewcommand{\arraystretch}{1}
\footnotesize
\caption{Performance Comparison Across Different Datasets}
\label{table1}
\begin{tabular}{llccccc}
\toprule
Datasets                              & Models            & ROUGE-1 & METEOR & BERTScore &GREEN \\
\midrule
\multirow{5}{*}{AbdomenAltas}         & LaMed-Phi-3-4B   & 0.136  & 0.058 & 0.807 & 0.011    \\
                                      & LaMed-Llama-2-7B & 0.139  & 0.060 & 0.810 & 0.009    \\
                                      & RadFM-14B        & 0.037  &0.013  & 0.794 & 0.000    \\
                                      & RadGPT-N &0.247 &0.112 &-  &-  \\
                                      & $\mu^2$LLM-1B(SFT)             &  0.529  & 0.295 & 0.891 & 0.281          \\
                                      & $\mu^2$LLM-1B(SFT\&DPO)             &  \textbf{0.567}  & \textbf{0.319} & \textbf{0.895} & \textbf{0.346}          \\
\midrule
\multirow{5}{*}{CT-Rate}              & LaMed-Phi-3-4B   &0.130   &0.050  & 0.814 & 0.002    \\
                                      & LaMed-Llama-2-7B & 0.103  & 0.048 & 0.815 & 0.001    \\
                                      & RadFM-14B        & 0.054  & 0.017 & 0.812 & 0.014    \\
                                      & CT-CHAT-8B &0.294&0.221&0.815&0.113 \\
                                      & $\mu^2$LLM-1B(SFT)             &0.517  &0.330 &0.879 &0.384          \\
                                      & $\mu^2$LLM-1B(SFT\&DPO)             &\textbf{0.539}  & \textbf{0.359} & \textbf{0.890} & \textbf{0.429}          \\
\midrule
\multirow{5}{*}{AMOS-MM}              & LaMed-Phi-3-4B   & 0.126  & 0.047 & 0.821 & 0.009    \\
                                      & LaMed-Llama-2-7B & 0.163  & 0.065 & 0.823 & 0.009    \\
                                      & RadFM-14B        & 0.046  & 0.015 & 0.812 & 0.001    \\
                                      & $\mu^2$LLM-1B(SFT)             & 0.421  & 0.249 & 0.881 & 0.339    \\
                                      & $\mu^2$LLM-1B(SFT\&DPO)             & \textbf{0.459}  & \textbf{0.876} & \textbf{0.881} & \textbf{0.400}    \\
\bottomrule
\end{tabular}
\end{table}

\begin{table}[!h]
\centering
\setlength{\tabcolsep}{1mm}
\renewcommand{\arraystretch}{1}
\footnotesize
\caption{Performance Comparison component effectiveness of $\mu^2$Tokenizer}\label{tabel2}
\begin{tabular}{llccccc}
\toprule
Model   & BLEU & ROUGE-1 & METEOR & BERTScore& GREEN \\
\midrule
Baseline       & 0.190             & 0.405                & 0.210               & 0.864 & 0.204                  \\

+RPE  & 0.281             & 0.421                & 0.236               & 0.880 & 0.277                 \\

+DTS & 0.271             & 0.411                & 0.240               & 0.888 & 0.299                 \\

+DMTP  & 0.254             & 0.401                & 0.220               & 0.874 & 0.233                   \\
$\mu^2$LLM-1B(SFT)  &0.279& 0.421  & 0.249 & 0.881 & 0.339   \\
$\mu^2$LLM-1B(SFT\&DPO)        &0.336 & 0.459  & 0.876 & 0.881 & 0.400   \\
\bottomrule
\end{tabular}
\end{table}

\textbf{Results Analysis.} 
Our model demonstrates superior performance compared to baseline models across multiple datasets. Notably, despite its significantly smaller scale (1B parameters, only 14\% of comparable models ranging from 7B to 14B), our model consistently outperforms these larger counterparts. Table~\ref{table1} presents a comparative evaluation across different datasets. Our model achieves state-of-the-art results, outperforming existing approaches. For instance, on the CT-Rate dataset, our model attains ROUGE-1 = 0.517, METEOR = 0.330, BERTScore = 0.879, and GREEN = 0.384, significantly surpassing CT-CHAT-8B. These results underscore the effectiveness of our approach, particularly after SFT and DPO. The integration of the GREEN Score-based dataset selection for DPO fine-tuning further enhances model performance, leading to more accurate and clinically relevant report generation. Specifically, the GREEN Score evaluation indicates a model capability improvement of 20\% with DPO in GREEN Score, with the most substantial gain observed on the AMOS-MM dataset, where the GREEN Score increased from 0.33 to 0.40. To further assess the impact of individual components, we conducted ablation experiments (Table~\ref{tabel2}). These experiments confirm that each enhancement contributes meaningfully to model performance. When training parameters remain consistent, incorporating DTS yields the most substantial performance boost, improving the GREEN Score by up to 0.2 points. This finding highlights the effectiveness of DTS in optimizing token representation and selection, leading to more accurate and clinically meaningful text generation.

\begin{figure}[!t]
    \centering
    \includegraphics[width=\textwidth]{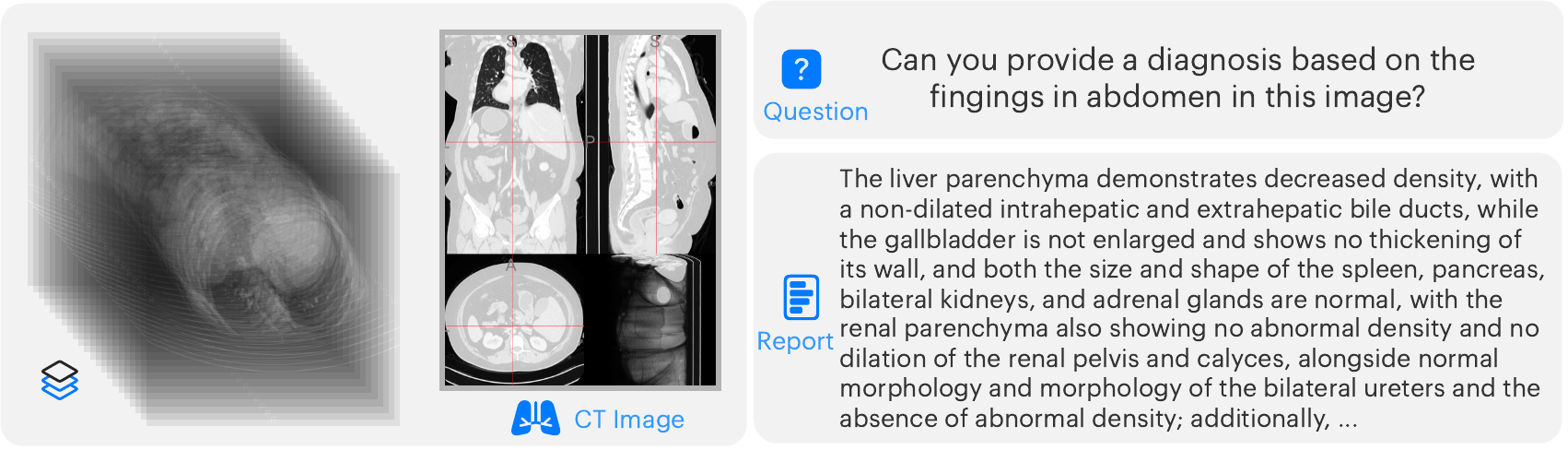}
    \caption{An example of the generated report from our $\mu^2$LLM-1B (SFT\&DPO).}
    \label{example}
\end{figure}

Figure~\ref{example} illustrates an example report produced by $\mu^2$LLM-1B. On the left, a 3D heat map visualizes the Question-CT cross-attention scores, indicating the regions of the CT scan most relevant to the model’s diagnostic reasoning. The center and right images depict the original CT scan and corresponding problem statement, where the model is tasked with identifying and diagnosing abnormalities in the abdominal region. The rightmost section displays the generated radiology report, which provides a structured interpretation of the CT findings. The report includes descriptions of liver parenchyma density, gallbladder morphology, renal pelvis dilation, and other critical observations. The generated text demonstrates clinical coherence and diagnostic accuracy, aligning with standard radiology interpretations. This visualization highlights the model’s capability to focus on diagnostically relevant regions and produce detailed, structured radiology reports, supporting its potential use in automated medical imaging analysis.

\section{Conclusion}

In this study, we introduced $\mu^2$Tokenizer, a multi-scale, multi-modal middleware, and a DPO optimization framework for radiology report generation. By integrating ViT3D with an LLM, our approach effectively combines visual and textual information, enabling accurate and coherent medical reports. To refine the model for RRG tasks, we utilized the GREEN-Model and SFT to curate high-quality datasets for DPO fine-tuning, improving alignment with clinical standards. Despite limited training data, our model outperformed larger baselines, particularly in GREEN Score, demonstrating the effectiveness of multi-modal fusion and optimization techniques in automated radiology reporting. By jointly supplying questions with large-scale LLM, answers and anatomically ordered chains-of-thought, the framework lowers annotation costs and paves the way for trustworthy, clinically grounded VQA models in medical imaging. These results highlight the importance of structured fine-tuning in enhancing diagnostic accuracy. Moving forward, our approach could be further extended to other medical imaging modalities and clinical applications.

\bibliographystyle{splncs04}
\bibliography{ref}

\begin{thebibliography}{10}
\providecommand{\url}[1]{\texttt{#1}}
\providecommand{\urlprefix}{URL }
\providecommand{\doi}[1]{https://doi.org/#1}

\bibitem{bai2024m3d}
Bai, F., Du, Y., Huang, T., Meng, M.Q.H., Zhao, B.: M3d: Advancing 3d medical image analysis with multi-modal large language models (2024)

\bibitem{banerjee-lavie-2005-meteor}
Banerjee, S., Lavie, A.: {METEOR}: An automatic metric for {MT} evaluation with improved correlation with human judgments. In: Goldstein, J., Lavie, A., Lin, C.Y., Voss, C. (eds.) Proceedings of the {ACL} Workshop on Intrinsic and Extrinsic Evaluation Measures for Machine Translation and/or Summarization. pp. 65--72. Association for Computational Linguistics, Ann Arbor, Michigan (Jun 2005)

\bibitem{bassi2025radgpt}
Bassi, P.R., Yavuz, M.C., Wang, K., Chen, X., Li, W., Decherchi, S., Cavalli, A., Yang, Y., Yuille, A., Zhou, Z.: Radgpt: Constructing 3d image-text tumor datasets. arXiv preprint arXiv:2501.04678  (2025)

\bibitem{everlight_radiology_2025}
{Everlight}: Radiology unlocked: The global radiologist report 2025

\bibitem{gao2024linvt}
Gao, L., Zhong, Y., Zeng, Y., Tan, H., Li, D., Zhao, Z.: Linvt: Empower your image-level large language model to understand videos. arXiv preprint arXiv:2412.05185  (2024)

\bibitem{hamamci2024developinggeneralistfoundationmodels}
Hamamci, I.E., Er, S., Almas, F., Simsek, A.G., Esirgun, S.N., Dogan, I., Dasdelen, M.F., Durugol, O.F., Wittmann, B., Amiranashvili, T., Simsar, E., Simsar, M., Erdemir, E.B., Alanbay, A., Sekuboyina, A., Lafci, B., Bluethgen, C., Ozdemir, M.K., Menze, B.: Developing generalist foundation models from a multimodal dataset for 3d computed tomography (2024)

\bibitem{hamamci2024foundation}
Hamamci, I.E., Er, S., Almas, F., Simsek, A.G., Esirgun, S.N., Dogan, I., Dasdelen, M.F., Wittmann, B., Simsar, E., Simsar, M., et~al.: A foundation model utilizing chest ct volumes and radiology reports for supervised-level zero-shot detection of abnormalities. arXiv preprint arXiv:2403.17834  (2024)

\bibitem{jaegle2021perceiver}
Jaegle, A., Gimeno, F., Brock, A., Vinyals, O., Zisserman, A., Carreira, J.: Perceiver: General perception with iterative attention. In: International conference on machine learning. pp. 4651--4664. PMLR (2021)

\bibitem{ji2022amos}
Ji, Y., Bai, H., Ge, C., Yang, J., Zhu, Y., Zhang, R., Li, Z., Zhanng, L., Ma, W., Wan, X., et~al.: Amos: A large-scale abdominal multi-organ benchmark for versatile medical image segmentation. Advances in Neural Information Processing Systems  \textbf{35},  36722--36732 (2022)

\bibitem{lin-2004-rouge}
Lin, C.Y.: {ROUGE}: A package for automatic evaluation of summaries. In: Text Summarization Branches Out. pp. 74--81. Association for Computational Linguistics, Barcelona, Spain (Jul 2004)

\bibitem{liu2023llava}
Liu, H., Li, C., Wu, Q., Lee, Y.J.: Visual instruction tuning (2023)

\bibitem{ostmeier2024greengenerativeradiologyreport}
Ostmeier, S., Xu, J., Chen, Z., Varma, M., Blankemeier, L., Bluethgen, C., Md, A.E.M., Moseley, M., Langlotz, C., Chaudhari, A.S., Delbrouck, J.B.: {GREEN}: Generative radiology report evaluation and error notation. In: Al-Onaizan, Y., Bansal, M., Chen, Y.N. (eds.) Findings of the Association for Computational Linguistics: EMNLP 2024. pp. 374--390. Association for Computational Linguistics, Miami, Florida, USA (Nov 2024). \doi{10.18653/v1/2024.findings-emnlp.21}

\bibitem{papineni-etal-2002-bleu}
Papineni, K., Roukos, S., Ward, T., Zhu, W.J.: {B}leu: a method for automatic evaluation of machine translation. In: Isabelle, P., Charniak, E., Lin, D. (eds.) Proceedings of the 40th Annual Meeting of the Association for Computational Linguistics. pp. 311--318. Association for Computational Linguistics, Philadelphia, Pennsylvania, USA (Jul 2002). \doi{10.3115/1073083.1073135}

\bibitem{rafailov2023direct}
Rafailov, R., Sharma, A., Mitchell, E., Manning, C.D., Ermon, S., Finn, C.: Direct preference optimization: Your language model is secretly a reward model. In: Thirty-seventh Conference on Neural Information Processing Systems (2023)

\bibitem{rosenkrantz_differences_2017}
Rosenkrantz, A.B.: Differences in perceptions among radiologists, referring physicians, and patients regarding language for incidental findings reporting  \textbf{208}(1),  140--143. \doi{10.2214/AJR.16.16633}, publisher: American Roentgen Ray Society

\bibitem{shaw2018selfattentionrelativepositionrepresentations}
Shaw, P., Uszkoreit, J., Vaswani, A.: Self-attention with relative position representations (2018)

\bibitem{thirunavukarasu_large_2023}
Thirunavukarasu, A.J., Ting, D.S.J., Elangovan, K., Gutierrez, L., Tan, T.F., Ting, D.S.W.: Large language models in medicine  \textbf{29},  1930--1940. \doi{10.1038/s41591-023-02448-8}

\bibitem{touvron2023llamaopenefficientfoundation}
Touvron, H., Lavril, T., Izacard, G., Martinet, X., Lachaux, M.A., Lacroix, T., Rozière, B., Goyal, N., Hambro, E., Azhar, F., Rodriguez, A., Joulin, A., Grave, E., Lample, G.: Llama: Open and efficient foundation language models (2023)

\bibitem{wu2023generalist}
Wu, C., Zhang, X., Zhang, Y., Wang, Y., Xie, W.: Towards generalist foundation model for radiology (2023)

\bibitem{bert_score}
Zhang, T., Kishore, V., Wu, F., Weinberger, K.Q., Artzi, Y.: Bertscore: Evaluating text generation with {BERT}. CoRR  \textbf{abs/1904.09675} (2019)

\bibitem{zhao2023radiology}
Zhao, G., Zhao, Z., Gong, W., Li, F.: Radiology report generation with medical knowledge and multilevel image-report alignment: A new method and its verification. Artificial Intelligence in Medicine  \textbf{146},  102714 (2023)

\end{thebibliography}

\newpage

\appendix
\section{Appendix}
\subsection{CT Report Rewriting: Prompt}
\begin{lstlisting}
You are an expert radiologists. And your task is to paraphrase a given radiology report. 
You need to:
1. Take the following 3 examples for style of writing.
2. You MUST NOT change any meaning of the original report, nor add or remove any information, not event correction.
3. Give out the paraphrased report directly, without any other content.
4. In English only.

Here are some examples of CT reports:
{SOME EXAMPLES OF DATASETS}

The original report:
```
{}
```
\end{lstlisting}
\subsection{CT Report Reasoning Synthesis: Prompt}
\subsubsection{1. Questions generation}
Use the following prompt to call LLM to generate a question list, and then use the regular expression \verb|r".*?\d\. ?([\^\n]*)"| to extract the questions separately:
\begin{lstlisting}
Here is a medical radiology report for a CT image.
```
{report}
```

Imagine you are assessing a student who is looking at a CT image, you are going to ask a list of questions. Don't mention the report, just list out as the form of questions, and questions only, in sequenced list.
\end{lstlisting}

\subsubsection{2. Answer\&Thinking generation}
Use the following prompt to call LLM to generate the thinking process and answers, and extract the thought content with the regular expression \verb|r"Thinking: ?([^\n]*)"| and extract the answer content with the regular expression \verb|r"Answer: ?([^\n]*)"|:

\begin{lstlisting}
You are a radiology medicine expert.
Your task is to answer the following radiology medicine question, using the patient's medical record report provided below.
When writing your thought process, imagine you are directly reviewing the patient's radiology images (do not mention the report), and describe your logical reasoning step by step as an expert would.
Then, provide your final, correct answer to the question.
Your response will be used to guide and improve the training of a multimodal large language model for radiology medicine images.
And here is the radiology report that you can see:
```
{report}
```

Now we have a question:
```
{question}
```

Please consider and answer the question in the following format:

Thinking: <thought process>

Answer: <answer to the question>
\end{lstlisting}
\subsubsection{3. Filter}
Use the following prompt to call LLM to filter out incorrect questions or answers that were generated in the previous step:
\begin{lstlisting}
You are an expert in radiology. Now you are reviewing a some questions and answers made by another expert.
You need to determine if the question is proper for a radiology exam, and the answer is correct.

If the question is proper for a radiology exam, and the answer is correct, return "Yes".
If the question is not proper for a radiology exam, or the answer is incorrect, return "No".
Do not return anything else.

The Report:
```
{report}
```
Question: {question}
Answer: {answer}
\end{lstlisting}
\subsubsection{4. Refine Thinking}
Use the following prompt to call LLM to rewrite the thinking data generated in the previous step to make it more in line with VQA habits:
\begin{lstlisting}
Help me edit the narrative below:
- If the narrative refers to a report, you change it as if you see it from the radiology image
- Edit only the places mentioned above, leave all other text the same 
- Do not add/remove/change any other information
- Directly output the result text

**The narrative:**
```
{report}
```
\end{lstlisting}
\subsubsection{5. Report Thinking Synthesis}
Use the following prompt to call LLM to generate the data that generates thoughts in the report process. \verb|Thinking_before| is spliced using all the questions, thinking, and answers in the QA dataset:
\begin{lstlisting}
ou are a radiology medicine expert. Now you are looking at a radiology image.
Here is your self talk when viewing the image:
```
{thinking_before}
```

Please paraphrase the self talk text and output it as **thinking progress**. Remember:
- Do not add/remove/alter any information
- Mind the coherence and fluence of output
- Deductions are prefered
- Directly output the result text

Your output:
\end{lstlisting}
\subsection{CT Report Translation: Prompt}
\begin{lstlisting}
This is an {source_lang} to {target_lang} translation, please provide the {target_lang} translation for this text. \
Do not provide any explanations or text apart from the translation.
{source_lang}: {source_input}
\end{lstlisting}
\end{document}